\documentclass[lettersize,journal]{IEEEtran}
\usepackage{amsmath,amsfonts}
\usepackage{algorithmic}
\usepackage{algorithm}
\usepackage{array}
\usepackage[caption=false,font=normalsize,labelfont=sf,textfont=sf]{subfig}
\usepackage{textcomp}
\usepackage{stfloats}
\usepackage{url}
\usepackage{verbatim}
\usepackage{graphicx}
\usepackage{cite}
\usepackage{multirow}
\usepackage{color}
\usepackage{booktabs}
\usepackage[table]{xcolor}
\usepackage{subfloat}
\usepackage{hyperref}
\hypersetup{hidelinks}
\usepackage{listings}%
\usepackage[switch]{lineno}
\usepackage{cite}
\usepackage{colortbl}

\begin{document}

\title{Nighttime Person Re-Identification via Collaborative Enhancement Network with Multi-domain Learning}

\author{Andong Lu, Chenglong Li*, Tianrui Zha, Xiaofeng Wang, Jin Tang and Bin Luo*

\thanks{

    This work was supported in part by the National Natural Science Foundation of China (No. 62406002 and No. 62376004); and the China Postdoctoral Science Foundation (2024M760011); and the Anhui Province Science Fundation for Distinguished Young Scholars (No. 2208085J18).
    
    A. Lu, J. Tang and B. Luo are with Anhui Provincial Key Laboratory of Multimodal Cognitive Computation, School of Computer Science and Technology, Anhui University, Hefei 230601, China. 
    (e-mail: adlu\_ah@foxmail.com;  tangjin@ahu.edu.cn; luobin@ahu.edu.cn)

    T. Zha is with Vrije Universiteit Amsterdam, the Netherlands. (e-mail: robinsonzha@gmail.com)
    
    C. Li is with Information Materials and Intelligent Sensing Laboratory of Anhui Province, Anhui Provincial Key Laboratory of Multimodal Cognitive Computation, School of Artificial Intelligence, Anhui University, Hefei 230601, China. 
    (e-mail: lcl1314@foxmail.com) 

    X. Wang is with School of Computer Science and Technology, Hefei University, Hefei 230601, China. (e-mail: xfwang@hfuu.edu.cn)
    
    (*Co-Corresponding author: Chenlong Li, Bin Luo)
    }      
    }
\markboth{Journal of \LaTeX\ Class Files,~Vol.~14, No.~8, August~2021}%
{Shell \MakeLowercase{\textit{et al.}}: A Sample Article Using IEEEtran.cls for IEEE Journals}


\maketitle
\begin{abstract}
Prevalent nighttime person re-identification (ReID) methods typically combine image relighting and ReID networks in a sequential manner. However, their performance (recognition accuracy) is limited by the quality of relighting images and insufficient collaboration between image relighting and ReID tasks. To handle these problems, we propose a novel Collaborative Enhancement Network called CENet, which performs the multilevel feature interactions in a parallel framework, for nighttime person ReID. In particular, the designed parallel structure of CENet can not only avoid the impact of the quality of relighting images on ReID performance, but also allow us to mine the collaborative relations between image relighting and person ReID tasks. To this end, we integrate the multilevel feature interactions in CENet, where we first share the Transformer encoder to build the low-level feature interaction, and then perform the feature distillation that transfers the high-level features from image relighting to ReID, thereby alleviating the severe image degradation issue caused by the nighttime scenario while avoiding the impact of relighting images. In addition, the sizes of existing real-world nighttime person ReID datasets are limited, and large-scale synthetic ones exhibit substantial domain gaps with real-world data. To leverage both small-scale real-world and large-scale synthetic training data, we develop a multi-domain learning algorithm, which alternately utilizes both kinds of data to reduce the inter-domain difference in training procedure. Extensive experiments on two real nighttime datasets, \textit{Night600} and \textit{RGBNT201$_{rgb}$}, and a synthetic nighttime ReID dataset are conducted to validate the effectiveness of CENet. We release the code and synthetic dataset at: \hyperlink{https://github.com/Alexadlu/CENet}{\color{red} https://github.com/Alexadlu/CENet}.
\end{abstract}


\section{Introduction}
\label{sec:intro}
\begin{figure}[t]
    \centering
    \begin{tabular}[b]{cc}
    \includegraphics[scale=0.45]
    {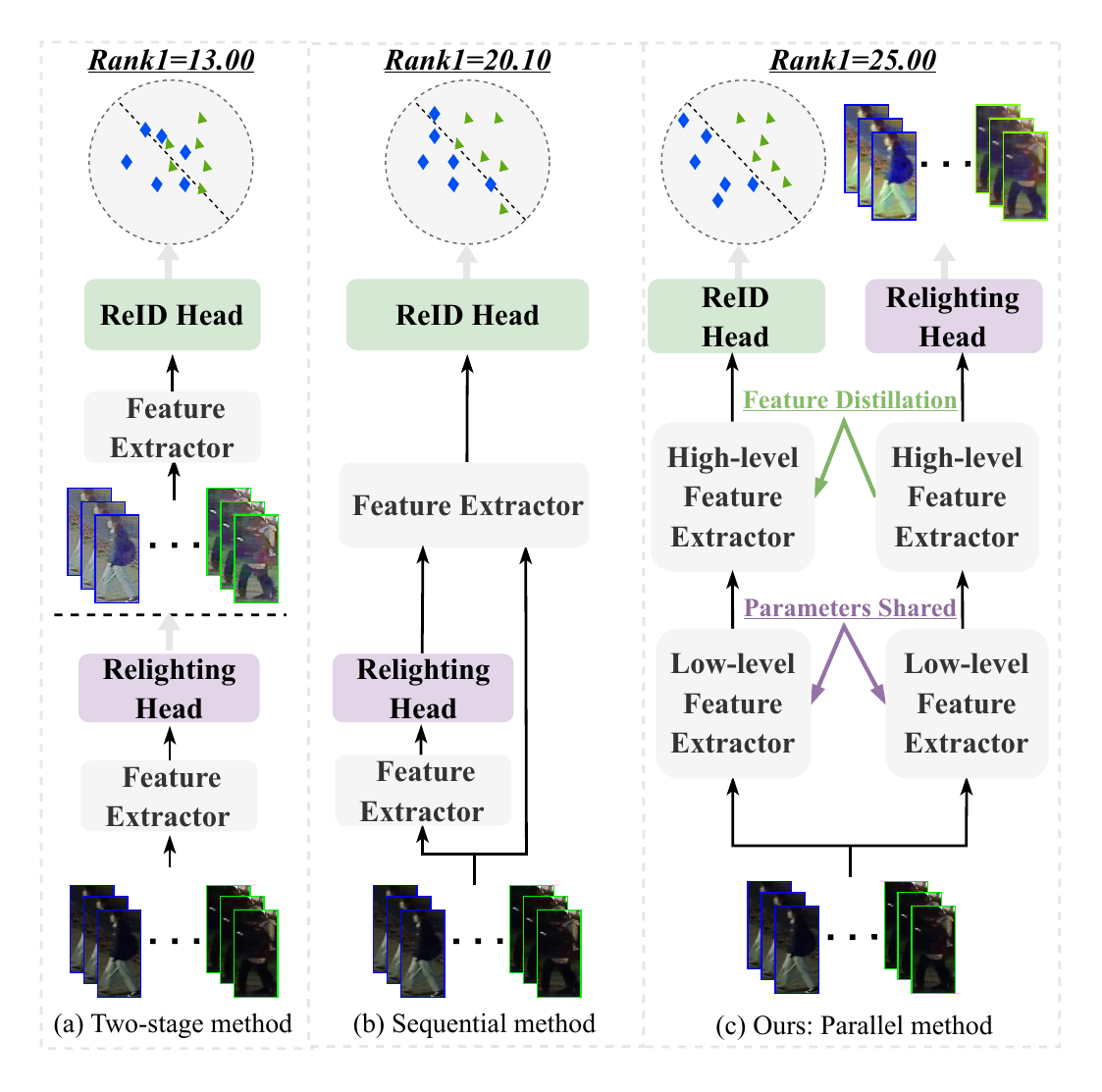}
    \end{tabular} 
    \caption{Comparison of three nighttime person ReID strategies. (a) uses Zero\_DCE~\cite{DCE_2020} for low-light image pre-processing and TransReID~\cite{TransReID_2021} for recognition. (b) represents the current state-of-the-art sequential end-to-end nighttime ReID method IDF~\cite{Lu2023night600} with. (c) introduces our proposed parallel end-to-end method, which outperforms the other strategies. (b) and (c) have consistent training strategy and training data to accurately reflect advantages of the parallel framework. The Rank values of the three strategies are evaluated on the same nighttime ReID dataset \textit{Night600}~\cite{Lu2023night600}}.
    \label{figure:motivation_fig}

\end{figure}

\IEEEPARstart{P}{erson} re-identification (ReID) is the task of identifying individuals in the images captured by different surveillance cameras, and it has significant potential applications in the fields of public security and criminal investigation. Despite current state-of-the-art methods~\cite{zhang_dscl_tifs,market1501_2015,MSMT17, TransReID_2021, DTRM_tifs} achieve excellent performance during the daytime, the challenges of person ReID in nighttime scenario remain unsolved. 
Specifically, unlike existing ReID studies that mainly focus on challenges such as pedestrian viewpoint differences and pedestrian occlusion, nighttime person ReID additionally introduces some challenges such as low light, imaging noise, and color shift caused by the limitations of imaging devices in nighttime scenarios. It raises significant challenges for existing ReID algorithms.
A straightforward approach to address the nighttime person ReID problem commonly involves the application of image relighting technique to enhance the visibility of input images before feeding them into ReID model, as shown in Figure~\ref{figure:motivation_fig} (a). However, these methods are not always effective because conventional relighting techniques are not specifically tailored to the ReID task. 

%
%


To overcome this limitation, existing works~\cite{Zhang2019NightPR, Lu2023night600} propose end-to-end nighttime person ReID methods that integrate image relighting and ReID within a single framework, as depicted in Figure~\ref{figure:motivation_fig} (b). For instance, Zhang et al.~\cite{Zhang2019NightPR} propose an end-to-end network that optimizes both the denoising and ReID objectives, leading to effective performance improvement compared to baseline method. Nevertheless, their approach overlooks the challenges posed by low-light conditions in nighttime images. Lu et al.~\cite{Lu2023night600} propose an illumination distillation network that combines features from nighttime images and relighting images to mitigate the limitations of solely relying on relighting images. Although these methods demonstrate promising performance improvements, they still follow a sequential framework of relighting and ReID networks. Consequently, the performance of ReID networks is not only constrained by the quality of relighting images, but also lacks effective collaborative modeling between these two tasks. Moreover, most existing relighting networks~\cite{DCE_2020,dce++tpami,zhang2020self} simply stack multiple convolutional layers, making it challenging to capture the crucial global context to handle lighting variations.


To address these challenges, we propose the Collaborative Enhancement Network (CENet) that integrates ReID and relighting networks in parallel, as depicted in Figure~\ref{figure:motivation_fig} (c). Such parallel structure design mitigates the impact of relighting image quality on ReID models, and also allows the multilevel feature interaction for effective collaborative modeling between ReID and relighting networks. 
%
%
The effective interaction between two heterogeneous tasks (relighting task and ReID task) is key to building collaborative enhancement. To this end, we propose a multilevel feature interaction scheme. In the low-level features, we adopt the current widely accepted shared-parameter design~\cite{wu2024asymmetric,he2024efficient} to achieve collaborative enhancement among low-level features. In high-level features, we innovatively design a feature distillation technique to achieve collaborative enhancement among high-level features while mitigating the impact on task-specific information.
By incorporating feature interactions at both low and high levels, we fully exploit the collaboration between these two tasks.
In particular, CENet comprises three primary components, including a shared encoder, a relighting subnet, and a ReID subnet. These components collaborate to learn lighting-enhanced features for both relighting and ReID tasks. Driven by relighting task, the shared encoder can extract lighting-enhanced features from nighttime images, which helps to enhance the low-level ReID features. In addition, to further utilize the lighting-enhanced features, a brightness feature distillation scheme is introduced from the relighting subnet to the ReID subnet to improve the high-level feature representations of ReID task. 

Constrained by the annotation difficulty of nighttime images, existing nighttime ReID datasets are small scale, which is hard to train big networks such as Transformer.
%
%
In this work, we synthesize a large-scale nighttime dataset called \textit{Syn\_Dark}, which is created by introducing some random factors including illumination, color, and contrast variations into two ReID datasets, \textit{Market1501}~\cite{market1501_2015} and \textit{MSMT17}~\cite{MSMT17}. 
Although introducing large-scale synthetic datasets is a straightforward idea, the gap between real and synthetic data makes the learning of large-scale synthetic data inadequate. To handle this issue, we propose a novel multi-domain learning algorithm which designs different learning strategies tailored to the characteristics of real and synthetic domains, respectively. It can effectively exploit large-scale synthetic data to improve the performance of nighttime person ReID task. In particular, we design tailored learning schemes for different domains, enabling us to leverage data from two distinct domains concurrently within a unified network. Through alternately optimizing the network parameters in different domain data, the network can achieve effective learning of domain-invariant features, thus mitigating inter-domain gaps. The feasibility of this scheme is validated in several domain adaptation studies~\cite{liu2023universal,li2024mdfl,royer2024scalarization}.
In addition, we introduce the multi-dimensional degradation strategy in synthetic data construction to improve the diversity of low-light scenes. We will release the synthetic dataset to public for free academic usage.

In summary, the main contributions are as follows.

\begin{itemize}

\item We propose a novel parallel framework to alleviate the negative impact and enhance the positive impact of the image relighting task on the person ReID task.  We release the code and synthetic dataset at: \hyperlink{https://github.com/mmic-lcl}{https://github.com/mmic-lcl}.

\item We design a new multilevel feature interaction scheme to facilitate collaborative modeling between the ReID and relighting networks. The shared low-level encoder aims to extract the collaborative features and the feature distillation in high-level features enhances the semantic information from relighting network to ReID network. 

\item We develop an effective multi-domain learning algorithm to utilize both advantages of large-scale synthetic and small-scale real data. The network optimization is alternately performed between synthetic and real domains, and their domain gap is thus alleviated in the training procedure.
\end{itemize}

The rest of this paper is organized as follows: Section II reviews related works. Section III introduces the methodology of CENet, including its architecture, multilevel feature interaction, and multi-domain learning. Section IV presents experimental results and comparisons. Finally, Section V concludes the paper.

\section{Related Work}
\label{sec:relatedwork}

\subsection{Person ReID Methods}

Due to challenges posed by low-light conditions, visible light cameras often struggle to capture clear images at night. Some studies~\cite{SYSU-MM01, nguyen2017person} propose to use visible light devices during the day and introduce infrared devices to capture person images, named cross-modality person ReID. 
However, cross-modality person ReID arises one of the major challenges is the modality gap between visible and infrared images. Therefore, many efforts~\cite{wang2019learning,HiCMD,XIVReID,LLCM,li2020infrared,vireid_trimodal_tifs,wang2020cross, zhang_dscl_tifs,Ye_2021_ICCV,du2023video,sun2024robust,DTRM_tifs} attempt to handle this issue. 
For example, Wang et al.\cite{wang2019learning} introduce a novel dual-level discrepancy reduction learning scheme, which handles the cross-modality discrepancies in image-level and feature-level, respectively. 
Choi et al.\cite{HiCMD} propose a hierarchical cross-modality disentanglement method to automatically disentangle ID-discriminative factors and ID-excluded factors from visible-thermal images. 
Li et al.\cite{XIVReID} design an auxiliary X modality and reformulate infrared-visible dual-modality cross-modality learning as an X-Infrared-Visible three-modality learning problem.
Zhang et al.\cite{LLCM} design a diverse embedding expansion network to effectively generate diverse embedding for nighttime cross-modality person ReID. Cui et al.\cite{DMA} adopt a dual-modality transfer module to perform compensation for information asymmetry in HSV color space, which can effectively alleviate cross-modality discrepancies and better preserve discriminative identity features.

Li et al.\cite{li2020infrared} employ a self-supervised learning network by establishing an intermediate modality to mitigate the modalities gap. Similarly, Ye et al.\cite{vireid_trimodal_tifs} reduce modality differences through an intermediate modality and employ a tri-modality joint learning strategy. To address the issue of potentially noisy labels in nighttime data annotations, Ye et al.\cite{DTRM_tifs} mine the cues at the channel-level, part-level, and graph-level. Zhang et al.\cite{zhang_dscl_tifs} develop a network to improve the consistency of features across different channels. Ye et al.\cite{Ye_2021_ICCV} propose a strategy involving random color channel swapping to create images that are not dependent on color within the same modality. 

Other studies~\cite{Ill_Invariant_2019ACMMM, Real_world_CVPR2020, IllAP_2020icip} synthesize low-light datasets to explore challenges of low-light images, but synthetic data are hard to reflect the real challenges of nighttime person ReID. To this end, Zhang et al.~\cite{Zhang2019NightPR} introduce a nighttime person ReID dataset focusing on the noise of nighttime images, and propose a denoising-then-match framework. Then, Lu et al.~\cite{Lu2023night600} propose a real nighttime ReID dataset that considers complex lighting conditions and develops an enhancement-then-match framework. However, these methods are still constrained by the quality of relighting images. In this work, we present a parallel architecture integrating relighting and ReID networks, which effectively avoids direct dependence on relighting images and also facilitates introducing the multilevel feature enhancement scheme between the two networks.

Moreover, the color shift issue, caused by imaging bias in clothes color under different low-light conditions, is also a challenge in nighttime person ReID. The challenge is actually similar to a challenge in clothing-change ReID. Existing studies on clothing-change ReID already achieve relatively mature progress.
For instance, Han et al.~\cite{han2023clothing} propose a clothing-change feature augmentation model that significantly expands the clothing-change data in feature space rather than visual image space.
Yang et al.~\cite{yang2023win} present an auxiliary-free competitive identification model, which achieves a win-win situation by enriching the identity-preserving information conveyed by the appearance and structure features while maintaining holistic efficiency.
This design might also benefit for nighttime person ReID. It is worth mentioning that nighttime data are difficult to annotate, and thus acquiring large-scale annotated nighttime person ReID data is extremely difficult. 
However, existing studies on unsupervised ReID provide some reference value for solving the learning problem of unlabelled data at night. For instance, Chen et al.~\cite{chen2023dual} propose a dual clustering co-teaching approach, which exploits a two-stream network to extract features and a dual clustering strategy to generate two sets of pseudo-labels for network training.

\subsection{Image Relighting Methods}


The image relighting task aims to recover invisible regions in low-light images. A classic method for adjusting image brightness is Histogram Equalization (HE)~\cite{dalal2005histograms}, which globally modifies the pixel distribution. However, it overlooks local information, potentially leading to loss of details, amplified noise, and suboptimal color reproduction. Some classic methods~\cite{dalal2005histograms,pizer1987adaptive,pisano1998contrast,kim1997contrast,land1977retinex,jobson1997properties} adjust image brightness by different theories and provide an important foundation for future research. Hence, various enhanced techniques~\cite{pizer1987adaptive,pisano1998contrast,kim1997contrast} are proposed to further improve enhancement results. Additionally, the \textit{RetineX} model~\cite{land1977retinex}, inspired by the human visual system, has gained widespread usage in enhancing low-light images. This model assumes that an image can be decomposed into an illumination map and a reflection map, with the reflection map, which represents the enhancement result.

\begin{figure*}[t]
     \centering
     \begin{tabular}[b]{cc}
     \includegraphics[scale=0.45]{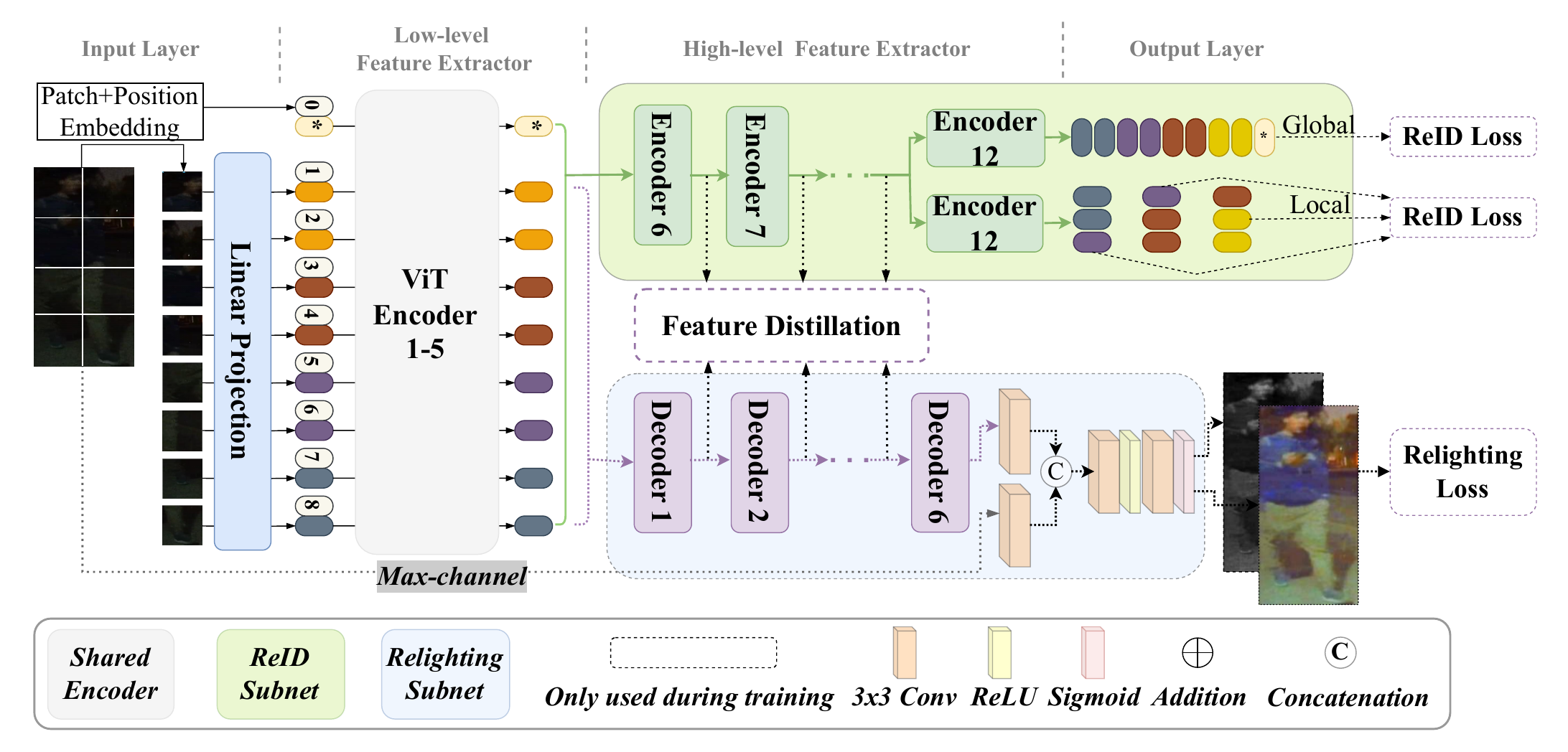}
     \end{tabular}
     \caption{The architecture of the Collaborative Enhancement Network (CENet) for nighttime person ReID. CENet consists of three parts: the shared encoder, the ReID subnet and the Relighting subnet. In addition, the dashed part will be removed during the inference phase.   }
     \label{figure:tcts_network_fig}
\end{figure*}

With the rise of deep learning, a series of methods combine convolutional neural networks and \textit{RetineX} theory~\cite{wei2018deep,zhang2021beyond,zhang2020self,kandula2023illumination,guo2023low} to achieve more accurate enhancement effects. Wei et al.~\cite{wei2018deep} first use RetineX theory with a deep neural network, employing a two-step approach of image decomposition followed by enhancement. Zhang et al.~\cite{zhang2021beyond} design a simple and effective network that maps the illumination and reflection maps to different subspaces, enabling separate corrections of brightness and degradation. However, these methods heavily rely on paired data for training, limiting their applicability. To address this limitation, Guo et al.~\cite{DCE_2020,dce++tpami} design a network to generate pixel-level curve parameter maps, which are then used to adjust low-light images. Zhang et al.~\cite{zhang2020self} propose an unsupervised relighting network based on information entropy and RetineX theory, which only uses low-light images for learning and outputs light and reflection maps. However, pretrained relighting networks cannot guarantee the optimal solution for nighttime person ReID.

In addition, we further discuss several methods that can perform global lighting in computer vision. Specifically, existing methods according to different learning strategies can be divided into two categories: supervised learning-based and unsupervised learning-based methods. Among them, supervised learning-based relighting methods~\cite{lu2020tbefn,yang2020fidelity} are current mainstream, but rely on large-scale paired data for training. However, it is extremely difficult to obtain paired data in real-world scenarios. Therefore, some studies~\cite{wang2020lightening} use synthetic techniques to build paired data for training. However, they do not yield satisfactory results due to the domain gap between real and synthetic data. In order to avoid paired data limitations, some researchers propose unsupervised relighting methods~\cite{wu2022uretinex,guo2023low}, and usually design various reference-free losses to guide model optimization. However, these methods focus on improving the quality of visual perception, and thus using them as a pre-processing step does not always guarantee the performance of nighttime person ReID. 

\section{Methodology}
\label{sec:method}

In this section, we introduce the proposed Collaborative Enhancement Network (CENet) in detail, including network overview, ReID branch, relighting branch, multilevel feature interactions, and multi-domain learning scheme.

\subsection{Architecture Overview}
The proposed framework, as shown in Figure~\ref{figure:tcts_network_fig}, comprises three main components, including the shared encoder, ReID subnet, and relighting subnet. 
%
%
We first utilize the shared encoder to extract features from nighttime person images. These features are then directed to two separate subnets of person ReID and image relighting.
The design of the shared encoder allows the ReID task to benefit from the low-level features relevant to the relighting task.
Then, we introduce a feature distillation mechanism to enhance the high-level feature representation of ReID task. 
%
In contrast to existing methods, our proposed network achieves effective collaborative modeling between image relighting and ReID tasks, while avoiding the impact of image lighting quality on ReID performance.

\subsection{ReID Branch}

The ReID branch adopts the Transformer network ViT~\cite{ViT_2021} with strong representation ability as the backbone network. Within this branch, the initial five encoders are designated as the shared encoder, while the remaining six encoders and task head serve as the ReID subnet. We partition the nighttime image $X$ into $N$ equally-sized image patches denoted as ${X_p^i|i=1,2,...,n}$, incorporating a trainable global token $X_{cls}$ as the input sequence for the ReID branch. Additionally, drawing from the TransReID~\cite{TransReID_2021}, we incorporate a learnable position embedding into the input sequence. Moreover, we introduce camera information embedding to enhance the acquisition of invariant features within the input sequences.
Subsequently, the input sequence is passed to the shared encoder, yielding the shared features, which can be expressed as:
\begin{equation}
  F_{share} = \left[\mathcal{E}\left(x_{cls}\right); \mathcal{E}\left(x_{p}^{1}\right); \mathcal{E}\left(x_{p} ^{2}\right) ; \cdots ; \mathcal{E}\left(x_{p}^{N}\right)\right],
\end{equation}
where $\mathcal{E}$ represents the shared encoder, and $F_{share}$ corresponds to the output features produced by the shared encoder.
%
Given that ReID is a classification task, extracting deep high-level semantic features is essential for distinguishing target individuals. To this end, we establish a ReID subnet comprising 6 Transformer encoders to extract more discriminative identity features. Finally, we employ a fully connected layer to adjust the number of extracted identity features for the final classification.

We employ the triplet loss and identity loss for joint optimization. 
In specific, the identity loss is the cross-entropy loss without label smoothing, which can be formulated as:
\begin{equation}
    \mathcal{L}_{ID}=-\frac{1}{B}\sum_{i=1}^{B}\sum_{j=1}^{N}y_{i,j}\ log{(\frac{e^{\alpha(s_{i,j}-m)}}{e^{\alpha(s_{i,j}-m)} + \sum_{k \neq j} e ^{\alpha s_{i,k}}})},
\end{equation}
where $B$ and $N$ represent the total numbers of samples and identities, respectively. Additionally, $y_{i,j}$ denotes the label assigned to the $j$-th identity of the $i$-th sample, while $s_{i,j}$ represents the score assigned to the $j$-th identity of the $i$-th sample. The variable $m$ represents the mean value of the identity score, and $\alpha$ is the temperature parameter.

The triplet loss is defined as follows:
\begin{equation}
   \mathcal{L}_{Tri}=\max(0, \alpha + d(f_{a}, f_{p}) - d(f_{a}, f_{n})),
\end{equation}
where $f_{a}$, $f_{p}$, and $f_{n}$ denote the feature representations of anchor samples, positive samples, and negative samples, respectively. Here, $d$ denotes the Euclidean distance, and $\alpha$ is a constant value used to control the feature margin. 
Furthermore, we optimize the ReID branch from both global and local perspectives, aiming to learn more fine-grained information. More detailed information can be referred to~\cite{TransReID_2021}.

\subsection{Relighting Branch}

As shown in Figure~\ref{figure:tcts_network_fig}, the relighting branch comprises a shared encoder and a relighting subnet. Considering that image relighting is an image reconstruction task rather than a classification task, we exclude the $X_{cls}$ component from $F_{share}$ and denote it as $F_{share}^*$. Subsequently, we forward it to the relighting subnet to perform image relighting.

\begin{figure}[h]
    \centering
    \begin{tabular}[b]{cc}
    \includegraphics[scale=0.45]{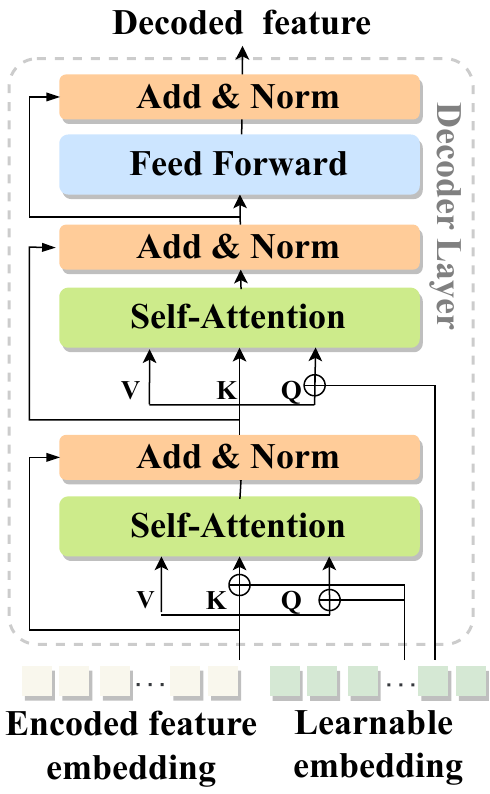}
    \end{tabular} 
    \caption{The illustration details the transformer decoder, where $\bigoplus$ represents feature summation.}
    \label{figure:decoder}
\end{figure}

Most of the existing deep relighting studies~\cite{DCE_2020,zhang2020self} are based on CNN network, which lacks the modeling ability of global lighting. To this end, we design a relighting branch based on Transformer with powerful global relationship modeling capabilities. The entire relighting branch contains a set of Transformer encoders fully shared with the ReID network and an independent relighting subnet.
The relighting subnet incorporates six Transformer decoders to decode the shared features $F_{share}^*$ and utilize them for subsequent image reconstruction. These decoders consist of multi-head self-attention modules and feedforward networks. 
Unlike the original Transformer decoder, which performs decoding iteratively, we decode the features of each position in parallel. Thus, within each decoder layer, the attention calculation uses the global representation $F_{share}^*$ and the task-specific learnable embedding $T_{e}$, which serve as position embeddings. As shown in Figure~\ref{figure:decoder}, the computations are defined as follows:
\begin{equation}
\begin{aligned}
& Q = W_{q1}(F_{share}^*+T_e),\\
& K = W_{k1}(F_{share}^*+T_e),\\
& V = W_{v1}(F_{share}^*),\\
& D_{out}^* = Norm(SA(Q,K,V)+F_{share}^*), \\
& Q^* = W_{q2}(D_{out}^*+T_e), \\
& K^* = W_{k2}(F_{share}^*), \\
& V^* = W_{v2}(F_{share}^*),\\
& D^{**}_{out} = Norm(SA(Q^*,K^*,V^*)+D_{out}^*),\\
& D_{out} = Norm(FFN(D^{**}_{out})+D^{**}_{out}),
\end{aligned}
\end{equation}
where $W_{q1},W_{k1},W_{v1},W_{q2},W_{k2},W_{v2}$ represent the projection layers. $Norm(.), SA(.)$ and $FFN(.)$ refer to layer normalization, self-attention layer, and feed-forward network in that order. $D_{out}$ denotes the output feature of the decoder layer. 

The output features of the decoder are further processed through a sequence of convolutional layers, combined with the maximum channel of the input image, to generate a reflectance map (relighting result) of the input image and a corresponding lighting map. Specifically, the output features of the decoder and the maximum channel of the input image undergo separate $3\times3$ convolutional operations for refinement. Subsequently, they are concatenated along the channel dimension and passed through two additional convolutional layers with activation functions to enable comprehensive fusion, resulting in a 4-channel image. Finally, the \textit{Sigmoid} function is applied to constrain these results. The first three channels represent reflection maps, while the last channel corresponds to the lighting map.
It is worth noting that we tailor different learning strategies for the relighting branch in both real and synthetic domain data. The details will be elaborated in Sec~\ref{mdl}.

\subsection{Multilevel Feature Interaction}

To improve ReID performance in low-light night scenes, we abandon the conventional approach~\cite{Lu2023night600, Zhang2019NightPR, night_segment} that extracts ReID features from relighting images. Instead, we focus on enhancing ReID feature representation from collaborative modeling between two tasks. Specifically, we adopt two strategies to achieve the enhancement from the relighting branch to the ReID branch. Since low-level networks typically capture common image features, we utilize parameter-sharing scheme between the ReID and relighting branches to improve low-level feature representation. Considering different tasks within high-level networks often model task-specific knowledge, which poses challenges for feature interaction. To handle this problem, we propose a feature distillation strategy to enhance high-level feature in the ReID branch.


\noindent\textbf{Low-level Shared Encoder.} The shared encoder is designed to extract crucial information from input nighttime images, serving both the joint learning ReID task and the relighting task. Unlike traditional convolutional neural networks with limited receptive fields, we equip with ViT encoders~\cite{ViT_2021}, which allows for better capturing of global information in nighttime images. Moreover, we exclude downsampling operations, thus preserving the spatial structure and detailed information in the shallower layers. This preservation is particularly advantageous for the image relighting process. Therefore, benefiting from the learning of the relighting task, the shared encoder will share the ability to extract lighting-enhanced features in both tasks. 

\noindent\textbf{High-level Feature Distillation.} High-level layers tend to model task-specific knowledge in various tasks. Therefore, it is a challenging problem to leverage the high-level features of the relighting subnet to enhance the high-level features of the ReID subnet. To address this issue, we adopt a non-intrusive interactive strategy known as feature distillation to enhance ReID features. By analyzing the correlation between the two tasks, 
we can conclude that the ReID task aims to improve the discriminative properties of foreground targets in an image, while the relighting task seeks to improve the global lighting level of an image. Therefore, there is some correlation between the two tasks. To achieve collaborative enhancement between the two tasks, we do not consider introducing an additional front-background separation technique for accurate distillation, instead utilizing a simple brightness loss function that is content-independent to boost the global lighting level of the ReID features. 

However, this overall brightness enhancement might amplify the brightness differences in person captured from various viewpoints. Consequently, we also incorporate a feature contrastive loss to ensure consistent feature representation for person from the same class, even under different lighting conditions. Finally, based on the above losses, we propose a novel feature distillation loss to achieve high-level feature interactions between relighting and Re-ID tasks and improve the performance of the ReID task. 
Specifically, the feature distillation loss consists of two items, the first of which aims to make the brightness features in ReID network approximate to the light-enhanced features. However, we observe that the performance improvement is limited when only using brightness loss distillation, as shown in Table~\ref{tab:tcte_ablation2}. We argue that although the enhancement of feature brightness can improve the feature representation ability, the risk of possible over-enhancement tends to decrease the differences between different individuals. Therefore, we introduce a contrast learning loss~\cite{khosla2020supervised} to pull together the features of persons in the same category with different brightnesses and push away the features of persons in different categories with similar brightness.
Overall, the feature distillation loss can be formulated as:
\begin{equation}
\label{eq12}
\mathcal{L}_{FD} = \underbrace{||\varphi (f_s), \varphi (f_t)||_2}_{\textit{brightness loss}} - \underbrace{log(\frac{exp(sim(f_s,f_t))}{\frac{1}{N} {\textstyle \sum_{i=1}^{N}exp(sim(f_s,f^i_t)} )})  }_{\textit{contrastive loss}}, 
\end{equation}
where $f_s$ and $f_t$ denote the high-level features of the two tasks, respectively. $\varphi()$ represents the feature channel mean function, and $||\cdot||_2$ denotes a mean square function. $N$ is the number of samples in the dataset, $\text{sim}(u, v)$ denotes a similarity measure between $u$ and $v$, and $i$ and $j$ are indices of data points, indicating whether $x_i$ and $x_j$ are samples from the same classes or different classes.

\subsection{Multi-Domain Learning}
\label{mdl}
\begin{figure}[t]
    \centering
    \begin{tabular}[b]{cc}
    \includegraphics[width=0.38\textwidth,height=8.5cm]{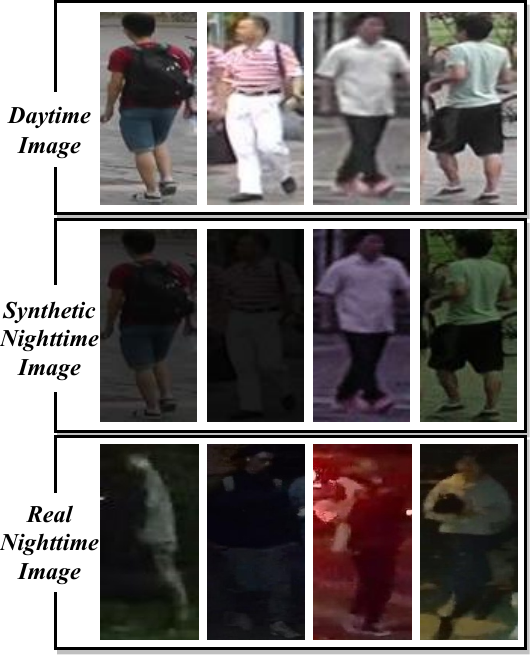}
    \end{tabular} 
    \caption{Comparison of real and synthetic domain data samples.}
    \label{figure:tcts_dark_imgs}
\end{figure}

\begin{figure}[t]
    \centering
    \begin{tabular}[b]{cc}
    \includegraphics[scale=0.38]{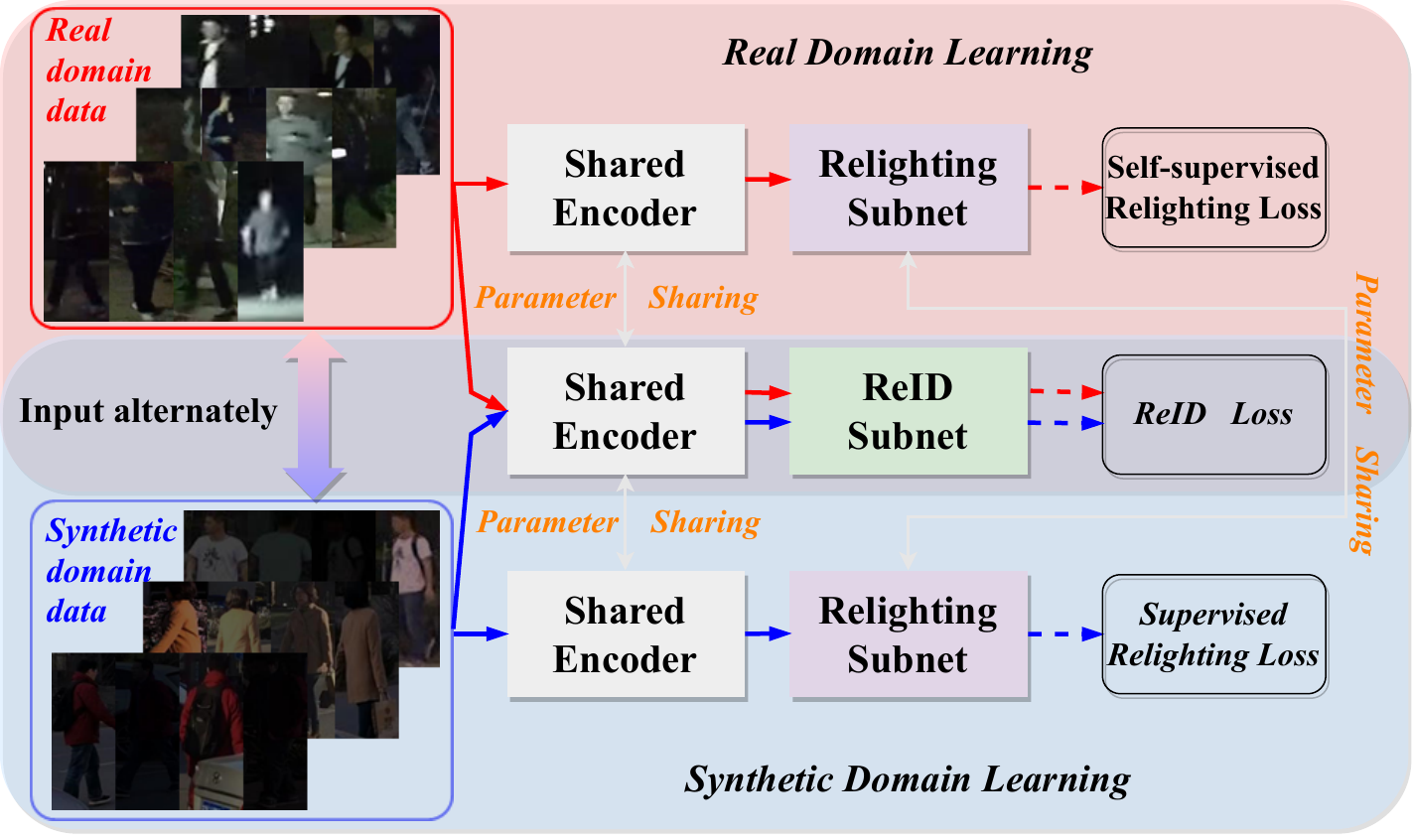}
    \end{tabular}
    \caption{Overview of the proposed multi-domain learning approach.}
    \label{figure:mutil_domain_learning}
    
\end{figure}

Constrained by the annotation difficulty of nighttime images, existing nighttime ReID datasets are small scale, which is hard to train big networks such as Transformer.
To overcome these challenges, we synthesize a large-scale nighttime person ReID dataset by applying multi-dimensional random degradation to existing daytime person ReID datasets (\textit{Market1501}~\cite{market1501_2015} and \textit{MSMT17}~\cite{MSMT17}). As depicted in Figure~\ref{figure:tcts_dark_imgs}, we apply lighting, contrast, color, and hue as four degradation factors to daytime images, generating diverse nighttime images that correspond to real nighttime scenarios. This approach offers two distinct advantages. First, the synthesized nighttime images maintain a one-to-one correspondence with their daytime counterparts, facilitating the establishment of supervised relighting learning strategies. Second, the introduction of synthetic data significantly enhances the diversity within the training set, thereby playing a crucial role in improving the model's generalization capabilities.
%
As shown in Figure~\ref{figure:mutil_domain_learning}, the ReID branch performs shared learning in the two domains, effectively enhancing its learning of domain-invariant knowledge. The learning process of the entire network alternates between the two domains, employing supervised relighting learning in the synthetic domain and unsupervised relighting learning in the real domain. In the following sections, we will provide a detailed explanation of the learning approaches adopted in both domains.

\noindent\textbf{Synthetic Domain Learning.}
In this stage, we establish a supervised low-light image relighting scheme utilizing synthetic low-light data and their corresponding well-lit counterparts. Given access to the corresponding ground truth, we employ the Euclidean distance as the loss function for this scheme, denoted as $\mathcal{L}_{IE_s}$, which can be expressed as:
\begin{equation}
\mathcal{L}_{IE_s} = \frac{1}{N} \sum_{i=1}^{N} \| I_{L_i} - I_{H_i} \|^2_2 ,
\end{equation}
where $N$ represents the total number of training samples, $I_{L_i}$ and $I_{H_i}$ denote the relighting result of the synthetic low-light image and corresponding well-lit image, respectively. $|| \cdot ||_2$ indicates a mean square function. This loss function quantifies the difference between the relighting result of the synthetic low-light image and its corresponding well-lit image, with the objective of minimizing this difference during training.

To facilitate joint optimization with the ReID branch, we introduce a hyperparameter $\lambda_1$ to balance the weights of the two loss functions. Consequently, the overall loss function at this stage can be expressed as:
\begin{equation}
\label{eq7}
    \mathcal{L}_{Syn} = \mathcal{L}_{ID} + \mathcal{L}_{Tri} + \lambda_1\mathcal{L}_{IE_s} + \lambda_2\mathcal{L}_{FD} ,
\end{equation}
where $\lambda_1 = 0.5$ and $\lambda_2 = 0.1$ serves as hyperparameters that control the weight of the relighting loss. 

\noindent\textbf{Real Domain Learning.} 
Given the challenges of obtaining highly corresponding low-light person images at night and well-lit person images during the day in real-world scenes, we design an unsupervised relighting scheme to train the relighting network in real domain data. Specifically, we employ the \textit{RetineX} model to decompose an image as follows:
\begin{equation}
    \mathcal{M} = \mathcal{R} \cdot \mathcal{I}.
\end{equation}
The \textit{RetineX} model provides a framework for separating the reflectance and lighting components in an image. In this case, the visible light image $\mathcal{M}$ is decomposed into the reflectance map $\mathcal{R}$, which captures the inherent characteristics of the scene, and the lighting map $\mathcal{I}$, representing the varying lighting conditions.

Relighting in unsupervised scenarios presents a highly ill-posed problem. To address this issue, we draw inspiration from~\cite{zhang2020self} and incorporate essential prior information along with four loss functions, including (1) reconstruction loss $\mathcal{L}_{rec}$, (2) reflection loss $\mathcal{L}_{ref}$, (3) color loss $\mathcal{L}_{col}$ (4) structure-aware smoothing loss $\mathcal{L}_{sa}$. 

First, we utilize the reconstruction loss ($\mathcal{L}_{rec}$) to preserve the invariance of image content and structural information. In this study, we employ a combination of mean absolute error loss and structural similarity loss for jointly constrain:
\begin{equation}
     \mathcal{L}_{rec} = 1-SSIM (\mathcal{R} \cdot \mathcal{I},\mathcal{M}) + \|\mathcal{R} \cdot \mathcal{I} - \mathcal{M} \|_{1} ,
\end{equation}
where ${SSIM(u,v)}$ and $\|\cdot\|_1$ represent the structural similarity index and $l_1$ norm, respectively.

Second, in low-light image relighting tasks, it is crucial to preserve maximum entropy in the enhanced image while maintaining consistency with the input image information. Specifically, the reflection map $\mathcal{R}$ is defined as the element-wise division of $\mathcal{M}$ by $\mathcal{I}$. Consequently, errors in estimating the lighting map can have exponential effects on the quality of the reflected image. To promote smoothness in the reflection map, a commonly employed regularization constraint is introduced:
\begin{equation}
\Delta \mathcal{R} = || \nabla_{x} \mathcal{R} ||_1 + || \nabla{y} \mathcal{R} ||_1 ,
\end{equation}
where $\Delta \mathcal{R}$ represents the difference between neighboring pixels in the reflection map, while $\nabla_{x}$ and $\nabla_{y}$ denote the gradients in the horizontal and vertical directions, respectively. Thus, the overall reflection loss can be defined as follows:
\begin{equation}
\mathcal{L}_{ref} = \left\|\max _{c \in r, g, b} R^{c}-\mathcal{H}\left(\max _{c \in r, g, b} M^{c}\right)\right\|_{1} + \Delta \mathcal{R},
\end{equation}
where $\mathcal{H}()$ denotes histogram equalization applied to the image, and $\max_{c \in r, g, b}$ selects the channel with the maximum pixel value in the image. 

Third, a color loss is introduced to mitigate color shifts during the enhancement process, leveraging the Gray-world color constancy hypothesis~\cite{chollet2017xception}, which can be described as follows:
\begin{equation}
\begin{aligned}
&\mathcal{L}_{col} =\sum_{\forall(i,j) \in \varepsilon}\left(R^{i}-R^{j}\right)^{2}, \varepsilon=\{(r, g),(r, b),(g, b)\},\\
\end{aligned}
\label{eq::5}
\end{equation}
where $C^{i}$ indicates the average intensity value of the $i$-th channel of the lighting-enhanced image.

Finally, a structure-aware smoothing loss~\cite{wei2018deep} is applied to constrain the reconstruction of enhanced images:
\begin{equation}
\mathcal{L}_{sa} = \|\Delta \mathcal{I} \cdot \exp(-\Delta \mathcal{R})\|_{1}.
\end{equation}
This loss essentially weights the original TV function~\cite{chen2010adaptiveTV} ($||\Delta \mathcal{I}||_{1}$) by the gradient of reflectance.

Hence, the unsupervised relighting loss function is defined as follows:
\begin{equation}
\label{eq11}
\mathcal{L}_{IE_r} = \lambda_a\mathcal{L}_{rec} + \lambda_b\mathcal{L}_{ref} + \lambda_c\mathcal{L}_{col} + \lambda_d\mathcal{L}_{sa}.
\end{equation}
In this study, we follow~\cite{zhang2020self} to set $\lambda_a$, $\lambda_b$, $\lambda_c$, and $\lambda_d$ to 1, 0.1, 0.2, and 0.1, respectively.

Finally, we employ the unsupervised relighting loss defined in Eq~(\ref{eq11}) and ReID loss to jointly optimize the whole network. Therefore, the overall loss function for real domain data is defined as:
\begin{equation}
\mathcal{L}_{Real} = \mathcal{L}_{ID} + \mathcal{L}_{Tri} + \lambda_1\mathcal{L}_{IE_r} + \lambda_2\mathcal{L}_{FD}.
\end{equation}
Here, $\lambda_1$ and $\lambda_2$ are consistent with the settings of $\lambda_1$ and $\lambda_2$ setting in Eq~(\ref{eq7}).

\section{Experiment}
\label{sec:exp}
In this section, we present the experimental setup and the comprehensive evaluation and comparison of the proposed method on three datasets, including two real nighttime ReID datasets and one synthetic ReID dataset. In addition, we perform an ablation study to justify the effectiveness of each component in the proposed method.

\subsection{Experiment Settings}
\noindent\textbf{Implementation Details.} In this work, we utilize the TransReID~\cite{TransReID_2021} network as the backbone of ReID branch. The backbone is equipped with an \textit{ImageNet} pre-trained model for feature extraction. Corresponding classification heads are designed for both the real domain dataset and the synthetic domain dataset, tailored to their respective number of categories. In relighting branch, the shared encoder follows the aforementioned parameters. However, the relighting subnet uses randomly initialized parameters, which allow the network to learn specific features for image relighting. 
The entire framework is trained end-to-end on the real domain dataset \textit{Night600} and the synthetic domain dataset \textit{Syn\_Dark}. Specifically, all images are resized to a size of $256 \times 128$, and random horizontal flipping, cropping, and random erasing techniques are applied during training to avoid overfitting. Each batch size is set to 64, and the initial learning rate is set to 0.008. We employ the stochastic gradient descent (SGD) optimizer with a momentum of 0.9 and weight decay of 1e-4. During the testing phase, CENet executes only the ReID branch, thereby eliminating the extra computational cost of image relighting in the inference stage.


\noindent\textbf{Datasets.} We utilize the \textbf{\textit{Night600}} dataset~\cite{Lu2023night600} as our real domain data. This dataset comprises a total of 28,813 nighttime images featuring 600 different individuals. The images are captured using 8 cameras and divided into a training set and a testing set. Specifically, the training set contains data from 300 individuals, while the testing set includes the remaining 300 individuals from the dataset. 


We create a synthetic ReID dataset called \textbf{\emph{Syn\_Dark}} by adjusting the brightness, contrast, hue, and color parameter ranges of images from \emph{Market1501}~\cite{market1501_2015} and \emph{MSMT17}~\cite{MSMT17} datasets.
To simulate lighting conditions and color variations in nighttime images, we leverage professional image editing software, Photoshop V23.3.2, to create synthetic data. We adjust the input images in four dimensions, including brightness, contrast, hue, and color.
To reduce brightness and contrast, we set the brightness parameter range between (10, 38) and the contrast parameter range between (7, 30). In addition, we set the image hue and color parameter ranges to (7, 30) and (20, 35), respectively.
By randomly selecting a specific value within a predefined range, we apply the corresponding degradation proportion to the image. For example, if the randomly chosen parameter for brightness is 10, the image's brightness will be reduced to 10\% of its original level. The primary challenges of real nighttime image are low light, low contrast, and color shifts that occur under different low-light conditions. Therefore, we perform the degradation in the above four dimensions to maximally simulate the real data. By leveraging the random combinations of degradation parameters and the availability of large-scale daytime data, we are able to obtain a broader range of challenging nighttime images.
Representative samples from both datasets are shown in Figure~\ref{figure:tcts_dark_imgs}, highlighting differences in sharpness and noise between the real and synthetic domains. 


\begin{table}[t]
\centering
\setlength{\tabcolsep}{1.7mm}
\caption{Statistics of real and synthetic domain datasets.}
\renewcommand{\tablename}{table}
\renewcommand\arraystretch{1.7}
\resizebox{0.95\columnwidth}{!}{
\begin{tabular}{cccc}
\hline
\textbf{\emph{Dataset}}   & \textbf{Train}       & \textbf{Query}    & \textbf{Gallery}   \\ \hline
\textbf{ \textit{Night600}}  & 300/14462   & 300/2180 & 300/14351 \\
\textbf{\emph{RGBNT201$_{\textbf{rgb}}$}}  & 171/3951  & 30/836 & 30/836\\
\textbf{Syn\_Dark} & 4852/138984 & 750/3368 & 750/13115 \\ \hline
\end{tabular}}
\label{tab:datasets_Statistics}
\end{table}

Finally, we introduce a multi-modal ReID dataset named \emph{RGBNT201}~\cite{RGBNT201_aaai}. The dataset comprises three highly aligned modalities: RGB, thermal infrared, and near-infrared. It focuses on capturing low-light scenes at night. To evaluate the performance of unimodal ReID methods in nighttime scenes, we only utilize RGB modality data from this dataset and call it \textbf{\emph{RGBNT201$_{\textbf{rgb}}$}}. The dataset consists of 201 identities with 4,787 images in total. Among them, 171 identities are used as the training set, and the remaining 30 identities constitute the test set.

\begin{table*}[t]
\setlength\tabcolsep{4pt}
\renewcommand\arraystretch{1.7}
\centering
\caption{Comparison results of CENet with state-of-the-art ReID methods on real and synthetic nighttime ReID datasets.}
\resizebox{2\columnwidth}{!}{
\begin{tabular}{@{}ccccccccccccc@{}}
\toprule
\textbf{Testing Dataset}      & \multicolumn{4}{c}{\textbf{ \textit{Night600}}~\cite{Lu2023night600}}                                 & \multicolumn{4}{c}{\textbf{\emph{RGBNT201$_{\textbf{rgb}}$}}~\cite{RGBNT201_aaai}}                                & \multicolumn{4}{c}{\textbf{\textit{Syn\_Dark}}}                                \\ 
\cmidrule(lr){1-1}
\cmidrule(lr){2-5}
\cmidrule(lr){6-9}
\cmidrule(lr){10-13}
\textbf{Evaluation Protocols} & \textbf{mAP}   & \textbf{Rank-1} & \textbf{Rank-5} & \textbf{Rank-10} & \textbf{mAP}   & \textbf{Rank-1} & \textbf{Rank-5} & \textbf{Rank-10} & \textbf{mAP}   & \textbf{Rank-1} & \textbf{Rank-5} & \textbf{Rank-10} \\ 
\cmidrule(lr){1-1}
\cmidrule(lr){2-5}
\cmidrule(lr){6-9}
\cmidrule(lr){10-13}
\textbf{IDE+}~\cite{IDE+_2019}                 & 3.38  & 7.75   & 20.09  & 27.43   & 19.44 & 17.23  & 30.38 & 37.44   & 68.31 & 87.50  & 95.67 & 97.33   \\
\textbf{PCB}~\cite{PCB_2018}                  & 6.09 & 12.84  & 26.47  & 35.14   & 17.88 & 14.35  & 28.83  & 35.89   & 71.31 & 88.99  & 95.93  & 97.65  \\
\textbf{ABD-Net}~\cite{ABDNet_2019}              & 7.23 & 14.36  & 29.59  & 40.00   & 19.04 & 15.67 & 26.32  & 35.89  & 41.06 & 65.20  & 74.20  & 82.42   \\
\textbf{BoT}~\cite{BoT_2019}                  & 5.40 & 11.20  & 22.60  & 30.60  & 21.90 & 19.30 & 32.80  & 43.50   & 56.10 & 81.00  & 92.30 & 94.90   \\
\textbf{AGW}~\cite{AGW_2021}                  & 6.10 & 12.50  & 24.40  & 30.90   & 23.70 & 21.10 & 37.20  & 49.30   & 73.40 & 90.60 & 96.60  & 98.10  \\
\textbf{TransReID}~\cite{TransReID_2021}            & 8.40  & 16.00  & 31.10 & 39.90   & 36.10 & 34.70  & 53.60 & 63.20  & 75.50 & 89.90  & 96.30  & 97.70   \\
\textbf{IDF}~\cite{Lu2023night600}                  & 9.20  & 17.20  & 34.40  & 43.80   & 38.00      & 38.40       & 53.70      & 63.20       & 75.10      & 88.70      & 95.50       & 97.20        \\ 
\textbf{CENet}               & \textbf{{13.30}} & \textbf{25.00}  & \textbf{43.00}  & \textbf{51.70}   & \textbf{55.30}      & \textbf{57.40}       & \textbf{73.60}       & \textbf{81.50}        & \textbf{79.20}      & \textbf{91.70}       & \textbf{97.00}       & \textbf{98.20}        \\ \bottomrule
\end{tabular}}
\label{tab:overall_compare}
\end{table*}


Table~\ref{tab:datasets_Statistics} provides a detailed overview of the two real-world datasets and one synthetic dataset. The \textit{Syn\_Dark} dataset offers a greater number of unique identities and images compared to both the \textit{Night600} and \textit{RGBNT201$_{\textbf{rgb}}$} datasets. 
In particular, we can briefly discuss the diversity of synthetic dataset, which is mainly reflected in two aspects. First, in terms of data scale, the number of person IDs and images in the training set is 16.7 times and 9.6 times that of the real nighttime dataset, respectively. Second, in terms of the lighting conditions of the synthetic data, the multi-dimensional degradation strategies are designed to greatly enrich the diversity of low light images in the synthetic dataset.
Subsequent experiments demonstrate that exposing the model to a wider range of identities and variations is crucial for training and improving performance.

\subsection{Comparison with State-of-the-art Methods}
In this section, we compare CENet with six state-of-the-art ReID algorithms, including IDE+~\cite{IDE+_2019}, PCB~\cite{PCB_2018}, ABD-Net~\cite{ABDNet_2019} , AGW~\cite{AGW_2021}, BoT~\cite{BoT_2019}, and TransReID~\cite{TransReID_2021} on \textit{Night600}, \textit{$RGBNT201_{rgb}$} and \textit{Syn\_Dark} datasets. To ensure fair evaluation, we retrain all methods using nighttime data. Additionally, we compare CENet with IDF~\cite{Lu2023night600}, which is a state-of-the-art ReID method designed for night scenes. From the results in Table \ref{tab:overall_compare}, CENet shows significant performance advantages overall comparison algorithms on both real and synthetic nighttime person ReID datasets.

\noindent{\bf Evaluation on \textit{Night600}.} From the results in Table \ref{tab:overall_compare}, it can be seen that CENet achieves significant improvements of 4.1\% on mAP and 7.8\% on Rank-1 compared to IDF~\cite{Lu2023night600} on \textit{Night600} dataset. 
Notably, IDF is specifically designed for nighttime ReID. As a typical sequential combination of relighting and ReID networks, IDF relies on the quality of relighting images in the inference phase, and is thus limited in accuracy and efficiency. In contrast, CENet removes relighting subnet during the inference phase and directly performs ReID task on nighttime images. Therefore, these results demonstrate the significant superiority of the proposed CENet in accuracy and inference efficiency against other methods.

In addition, compared with the existing conventional ReID algorithms, CENet outperforms TransReID and AGW by 4.9\%/9.0\% and 7.2\%/12.5\% on mAP/Rank-1, respectively. In particular, the TransReID method, serving as the baseline for our approach, maintains a network structure identical to CENet during the inference stage. Therefore, the above experimental results strongly suggest the significant advantages of CENet in the night ReID task.

\noindent{\bf Evaluation on \textit{RGBNT201$_{\bf rgb}$}.} We compare our method with the advanced ReID algorithm on another real nighttime ReID dataset \textit{$RGBNT201_{rgb}$}, and the results are shown in the Table \ref{tab:overall_compare}. Similarly, CENet still exhibits a clear performance advantage. Specifically, compared to the second best IDF algorithm on this dataset, CENet achieves a significant improvement of 17.3\% and 19.0\% on mAP and Rank-1, respectively. Compared with TransReID and AGW, CENet outperforms them by 19.2\%/26.7\% and 21.8\%/36.3\% on mAP/Rank-1 metrics respectively. Although all methods perform better on the \textit{$RGBNT201_{rgb}$} dataset than on the \textit{Night600} dataset, this improvement can be attributed to the smaller scale of the \textit{$RGBNT201_{rgb}$} test set. With only 30 identities, this dataset does not fully represent the complexity of nighttime conditions. In addition, by comparing the performance of two nighttime datasets of different sizes, we believe that further expanding the size of nighttime training data is the key to improving nighttime ReID performance.

\noindent\textbf{Evaluation on \textit{Syn\_Dark}.} In Table \ref{tab:overall_compare}, we show the experimental results of CENet and existing state of the art methods on the synthetic nighttime ReID dataset. It can be seen that compared with IDF and TransReID, CENet achieves 4.1\%/3.0\% and 3.7\%/1.8\% performance improvement on mAP and Rank-1, respectively, which indicates that the proposed method demonstrates effective modeling capability for synthetic domain data.  
Notably, IDF, which combines relighting network and ReID network in a sequential manner, performs lower on synthetic datasets than conventional ReID algorithms (TransReID), suggesting that sequential structures on synthetic domain data are difficult to avoid performance degradation due to unstable relighting results. On the other hand, it also reflects the difference between synthetic data and real data. Therefore, exploring low-light scene ReID only on the synthetic dataset is hard to reflect the performance of real nighttime scenarios. However, compared with daytime, the current nighttime ReID performance is still low, which indicates the limitations of existing night ReID methods.

\noindent\textbf{Comparison with multimodal ReID.} To deal with the nighttime ReID problem, one potential approach is to incorporate visual modalities that are less affected by low-light conditions, thus mitigating the problem of poor imaging quality in RGB low-light scenes. In this context, Zheng \emph{et al.}~\cite{RGBNT201_aaai} propose a multimodal dataset to explore night ReID, which attracts attention to the research of multimodal ReID algorithms. However, the popularization of multimodal monitoring devices and the registration of multimodal data require substantial costs, which limits the practical applications of multimodal ReID. To this end, we explore the potential of low-light RGB images. In Table~\ref{tab:compare_trimodal}, we show the experimental results of CENet and existing multimodal ReID methods. It can be seen that CENet, which only uses RGB data for training and evaluation, performs much better than the algorithms that use multi-modal data for training and evaluation. Specifically, CENet achieves 8.9\%/10.3\% and 16.8\%/18.5\% performance advantages over IEEE and PFNet on mAP and Rank-1, respectively. 

It is worth discussing that current multi-modal ReID methods primarily adopt CNN network to extract person features, which lack crucial global modeling capabilities. In contrast, CENet employs a powerful Transformer network for feature extraction and incorporates synthetic low-light data during training, significantly improving the model's representation and generalization abilities. Consequently, CENet can achieve superior performance using only the RGB modality in low-light conditions. In addition, several Transformer-based multi-modal ReID algorithms~\cite{zhang2024magic,wang2024heterogeneous} published after this work achieve more superior performance, which still demonstrates the effectiveness of multi-modal information.
In summary, this indicates that existing multimodal ReID algorithms have not fully exploited the potential of the RGB modality, and that ReID under nighttime RGB images has great potential.


\begin{table}[t]
\centering
\setlength\tabcolsep{4pt}
\setlength{\tabcolsep}{1.7mm}
\caption{Comparison results of CENet with state-of-the-art multi-modality methods on \textit{RGBNT201} dataset}
\renewcommand\arraystretch{1.7}
\resizebox{0.9\columnwidth}{!}{
\begin{tabular}{@{}ccccccc@{}}
\toprule
\textbf{Methods}   & \textbf{Modality}         & \textbf{mAP}   & \textbf{Rank-1} & \textbf{Rank-5} & \textbf{Rank-10} \\ 
\cmidrule(lr){1-1}
\cmidrule(lr){2-2}
\cmidrule(lr){3-6}
\textbf{HAMNet}~\cite{li2020multi} & \multirow{3}{*}{RGB+NI+TIR} & 27.70  & 26.30  & 41.50 & 51.70  \\
\textbf{PFNet}~\cite{RGBNT201_aaai} &                               & 38.50 & 38.90  & 52.00  &58.40    \\
\textbf{IEEE}~\cite{wang2022interact}   &                               & 46.40  & 47.10  & 58.50   & 64.20   \\ \midrule
\textbf{CENet} &         RGB                 &55.30 & 57.40  & 73.60  &81.50   \\ \bottomrule
\end{tabular} }
\label{tab:compare_trimodal}
\end{table}

\noindent\textbf{Comparison with cross-modality ReID.} To further validate the effectiveness of the proposed method, we compare it with current well-developed cross-modal ReID methods, including DEEN~\cite{LLCM}, DMA~\cite{DMA}, HiCMD~\cite{HiCMD}, and XIVReID~\cite{XIVReID}. We regard a copy of night600 as the NIR modality to make the cross-modal ReID algorithm executable.

\begin{table}[t]
\centering
\setlength\tabcolsep{4pt}
\setlength{\tabcolsep}{1.7mm}
\caption{Comparison results of CENet with state-of-the-art cross-modality methods on \textit{Night600} dataset}
\renewcommand\arraystretch{1.7}
\resizebox{0.9\columnwidth}{!}{
\begin{tabular}{@{}ccccccc@{}}
\toprule
\textbf{Methods}   & \textbf{Pub. info}         & \textbf{mAP}   & \textbf{Rank-1} & \textbf{Rank-5} & \textbf{Rank-10} \\ 
\cmidrule(lr){1-1}
\cmidrule(lr){2-2}
\cmidrule(lr){3-6}
\textbf{HiCMD}~\cite{HiCMD} &    CVPR 2020     & 2.94  & 7.66   & 14.77   &  19.27     \\
\textbf{XIVReID}~\cite{XIVReID} &   AAAI 2020     & 3.29 & 14.27  & 31.70  &40.37      \\
\textbf{DEEN}~\cite{LLCM} & CVPR 2023   &12.84      & 21.56 & 36.97  &45.69    \\
\textbf{DMA}~\cite{RGBNT201_aaai} &  TIFS 2024      & 12.75 & 20.31  & 35.63  &43.83    \\
\midrule
\textbf{CENet} &           Ours             & \textbf{{13.30}} & \textbf{25.00}  & \textbf{43.00}  & \textbf{51.70}   \\ \bottomrule
\end{tabular} }
\label{tab:compare_crossmodal}
\end{table}

The results are shown in Table~\ref{tab:compare_crossmodal}, and it can be seen that the current state-of-the-art cross-modal ReID algorithms, DMA and DEEN, both achieve notable performance in \textit{Night600}. The reason can be attributed to the fact that these algorithms not only focus on the problem of modal differences, but also design the feature extraction network carefully. It indicates that improving the feature extraction capability of the network is one of the key factors to improve the performance of ReID at night. Therefore, the main reason for the poor performance of XIVReID and HiCMD in \textit{Night600} can be attributed to the fact that these two algorithms focus only on solving the cross-modal differences and neglect the careful design of the feature extraction network.
Although advanced cross-modal methods achieve good performance, our method exceeds the mAP/Rank1 metrics by 0.55\%/4.69\% and 0.46\%/3.44\% compared to DMA and DEEN, respectively, a result that supports the superiority of the proposed method. In addition, we observe that the DMA and DEEN algorithms have better performance than existing studies of unimodal algorithms, such as TransReID, AGW, etc., which suggests that the existing studies of cross-modal ReID may be more important references for visible-based nighttime ReID.

\begin{table}[ht]
\setlength\tabcolsep{4pt}
\centering
\setlength{\tabcolsep}{1.7mm}
\caption{Ablation studies of CENet on \textit{Night600} and \textit{Syn\_Dark} datasets}
\renewcommand\arraystretch{1.7}
\resizebox{0.9\columnwidth}{!}{
\begin{tabular}{lcccccccccc}
\toprule
          & \multicolumn{3}{c}{\textbf{ \textit{Night600}}} & \multicolumn{3}{c}{\textbf{\textit{Syn\_Dark}}}  \\ 
        \cmidrule(lr){2-4}
        \cmidrule(lr){5-7}
          & \textbf{mAP}  & \textbf{Rank-1}   & \textbf{Rank-5}     & \textbf{mAP}    & \textbf{Rank-1}    & \textbf{Rank-5}      \\ \hline

\textbf{CENet}   &13.30 & 25.00 & 43.00  &79.20  & 91.70  & 97.00 \\
\textbf{CENet w/o MD}     & 9.50 & 19.20 & 33.50  & 77.30   & 91.00 & 96.60   \\
\textbf{CENet w/o MD MFI} & 8.40 & 16.00 & 31.10 & 75.50 & 89.90  & 96.30   \\ 
\hline
\textbf{CENet w/o MD MFI$_{FD}$}    & 8.80 & 16.10 & 31.30  & 79.80  & 91.90  & 97.20   \\
\textbf{CENet w/o MD MFI$_{PS}$}   & 8.90 & 17.20 & 31.50  & 76.20 & 90.00  & 96.20  \\

\bottomrule
\label{tab:tcte_ablation}
\end{tabular}}
\end{table}

\subsection{Ablation Study}


\noindent{\textbf{Effectiveness of components.}} We perform an in-depth evaluation of the key components of CENet on two different datasets, \textit{Night600} and \textit{Syn\_Dark}. The major components under investigation are Multilevel Feature Interaction (MFI) and Multi-Domain learning (MD). Initially, we remove MD from CENet and label this model as “CENet w/o MD”. As shown in Table~\ref{tab:tcte_ablation}, this leads to significant performance drops on the \textit{Night600} dataset, with mAP and Rank-1 decreasing by 3.8\% and 5.8\% respectively. On the \textit{Syn\_Dark} dataset, mAP and Rank-1 declined by 1.9\% and 0.7\% respectively. It shows that MD plays a key role in the model's learning in both the synthetic and real domains. 
Subsequently, we exclude both MD and MFI and label the model as “CENet w/o MD MFI”, which is our baseline method. From the table ~\ref{tab:tcte_ablation}, it can be observed that the model performance in \textit{Night600} mAP decreased by 4.9\%, Rank-1 by 9\%, while mAP is down 3.7\% and Rank-1 is down 1.8\% on the \textit{Syn\_Dark} dataset.
These results further demonstrate the effectiveness of the designed MFI for ReID networks.

In addition, to further analyze the importance of each component in the multistage feature interaction, we conduct experiments on two feature interaction strategies respectively. “CENet w/o MD MFI$_{FD}$” stands for removing the multi-domain learning strategy and the feature distillation scheme of the multilevel feature interaction, to verify the effectiveness of parameter sharing. As shown in Table~\ref{tab:tcte_ablation}, compared with the baseline method, “CENet w/o MD MFI$_{FD}$” improves mAP and Rank-1 by 0.4\%/0.1\% and 4.3\%/2\% on \textit{Night600} and \textit{Syn\_Dark} datasets, respectively, verifying the effectiveness of the low-level feature interaction. Similarly, "CENet w/o MD MFI $_{PS}$" stands for removing the multi-domain learning strategy and the parameter-sharing strategy of the multilevel feature interaction. Compared with the baseline method, the mAP/Rank-1 of the "CENet w/o MD MFI $_{PS}$" on \textit{Night600} and \textit{Syn\_Dark} datasets are increased by 0.5\%/ 1.2\% and 0.7\%/ 0.1\% respectively, indicating the effectiveness of high-level feature interaction. 
In conclusion, the combination of the two interactions results in a more significant improvement, that is, an increase of 1.1\%/3.2\% and 1.8\%/1.1\% in the mAP/Rank-1 metrics on \textit{Night600} and \textit{Syn\_Dark} datasets, respectively.



\begin{table}[ht]
\setlength\tabcolsep{4pt}
\centering
\setlength{\tabcolsep}{1.7mm}
\caption{Feature distillation (FD) loss ablation studies on the \textit{Night600} dataset}
\renewcommand\arraystretch{1.7}
\resizebox{0.9\columnwidth}{!}{
\begin{tabular}{ccccc}
\toprule
   \textbf{ Methods }     & \textbf{mAP}  & \textbf{Rank-1}   & \textbf{Rank-5}  & \textbf{Rank-10} \\   
                  \cmidrule(lr){1-1}
                  \cmidrule(lr){2-5}
CENet-FD w/o MD    & 9.50 & 19.20 & 33.50  & 41.90    \\
CENet-BL w/o MD    & 9.00 & 17.00 & 31.20  & 39.60    \\
CENet-MSE w/o MD     & 7.50 & 14.20 & 29.20  & 38.40  \\
CENet-KL w/o MD    & 7.90 & 13.40 & 28.70  & 37.40  \\
\bottomrule
\label{tab:tcte_ablation2}
\end{tabular}}
\end{table}

\noindent{\textbf{Effectiveness of distillation loss.}} To evaluate the effectiveness of the proposed feature distillation (FD) loss, we also investigate two commonly employed distillation losses: mean squared error (MSE) loss and Kullback-Leibler (KL) loss for feature distillation. To this end, we construct three different variants, namely “CENet-BL w/o MD”, “CENet-MSE w/o MD”, and “CENet-KL w/o MD”, which respectively indicate replacing FD with only brightness loss, MSE loss, and KL loss. From the results in Table~\ref{tab:tcte_ablation2}, it can be observed that these alternative losses do not bring performance improvement when applied to the real nighttime dataset \textit{Night600}. This demonstrates the effectiveness of the proposed feature distillation loss. 

\begin{table}[ht]
\setlength\tabcolsep{4pt}
\centering
\setlength{\tabcolsep}{1.9mm}
\caption{Ablation studies of multi-domain learning on \textit{Night600} and \textit{Syn\_Dark} datasets}
\renewcommand\arraystretch{1.8}
\resizebox{1\columnwidth}{!}{
\begin{tabular}{lcccccccccc}
\toprule
          & \multicolumn{3}{c}{\textbf{ \textit{Night600}}} & \multicolumn{3}{c}{\textbf{\textit{Syn\_Dark}}}  \\ 
        \cmidrule(lr){2-4}
        \cmidrule(lr){5-7}
          & \textbf{mAP}  & \textbf{Rank-1}   & \textbf{Rank-5}     & \textbf{mAP}    & \textbf{Rank-1}    & \textbf{Rank-5}      \\ \hline
\textbf{Single-domain learning} & 9.50 & 19.20 & 33.50  & 77.30   & 91.00 & 96.60    \\ 
\textbf{Mix-domain learning} & 10.50 & 22.50 & 36.40 & 72.10 & 87.70  & 95.50   \\ 
\textbf{Multi-domain learning}   &\textbf{13.30} & \textbf{25.00} & \textbf{43.00}  &\textbf{79.20}  &\textbf{91.70}  & \textbf{97.00} \\

\bottomrule
\label{tab:tcte_ablation_MDL}
\end{tabular}}
\end{table}

\noindent{\textbf{Effectiveness of multi-domain learning.}} To evaluate the effectiveness of the proposed multi-domain learning, we also investigate two commonly employed learning scheme: single-domain learning and mix-domain learning for network training. 
Specifically, we design a single-domain learning scheme that trains exclusively on the real-domain dataset, and a mixed-domain learning scheme that trains on a combined dataset of real and synthetic domains. The multi-domain learning approach exhibits clear advantages, as reflected in Table~\ref{tab:tcte_ablation_MDL}. It surpasses the single-domain learning approach by 3.8\% and 4.8\% in mAP and Rank-1 metrics on the real domain test set. Moreover, when compared to the mixed-domain learning approach, multi-domain learning strategy demonstrates superior performance with a 2.8\% and 2.5\% increase in mAP and Rank-1 metrics, respectively. These results further validate the effectiveness and superiority of the multi-domain learning strategy.


\begin{table}[ht]
\setlength\tabcolsep{4pt}
\centering
\setlength{\tabcolsep}{1.3mm}
\caption{Ablation study of the importance of parallel architecture}
\renewcommand\arraystretch{1.1}
\resizebox{\linewidth}{!}{
\begin{tabular}{cccccc}
\toprule
\multirow{2}{*}{}     & \multicolumn{2}{c}{Traning dataset}                   & \multicolumn{2}{c}{Testing dataset} & \multirow{2}{*}{Improvement} \\ \cmidrule(lr){2-3} \cmidrule(lr){4-5} 
                      & Night600                  & Syn\_dark                 & Rank-1            & mAP             &                              \\ \hline
Baseline              & \checkmark &                           & 16.00             & 8.40            & --                           \\
w/ Data               & \checkmark & \checkmark & 18.10             & 8.90            & 2.1/0.5                      \\
w/ Parallel           & \checkmark &                           & 19.20             & 9.50            & 3.2/1.1     \\
w/ Data Parallel      & \checkmark & \checkmark & 22.50             & 10.50           & 4.4/1.6                      \\
w/ Data Parallel MD   & \checkmark & \checkmark & 25.00             & 13.30           & 2.5/2.8                      \\ \hline
w/ Data Sequential    & \checkmark & \checkmark & 20.30             & 10.00           & 2.2/1.1                      \\
w/ Data Sequential MD & \checkmark & \checkmark & 20.10             & 10.50           & -0.2/0.5                     \\ \bottomrule
\end{tabular}}
\label{tab:PA_ablation}
\end{table}

\noindent{\textbf{Importance and necessity of parallel architecture.}}
Since extra data and multi-domain learning are integrated into the parallel architecture, it is difficult to clearly demonstrate their benefits. Therefore, we conduct a series of experiments to illustrate the core impact of the parallel architecture. Specifically, we compare the parallel architecture (Parallel), extra data (Data), and multi-domain learning (MD) in two perspectives. First, each component is introduced incrementally in the baseline approach. As shown in Table~\ref{tab:PA_ablation}, the “w/ Parallel” and “w/ Data Parallel” show the parallel architecture improves Rank-1/mAP by 3.2\%/1.1\% and 4.4\%/1.6\% over the baseline method TransReID before and after the introduction of extra data, respectively. The “w/ Data” indicates that introducing extra data in TransReID brings 2.1\%/0.5\% improvement over TransReID. The “w/ Data Parallel MD” indicates that improves 2.5\%/2.8\% on Rank-1/mAP in the parallel architecture. Second, we introduce extra data and multi-domain learning into the sequential framework (Sequential) to further demonstrate the advantages of parallel framework. As shown in the last two rows of Table~\ref{tab:PA_ablation}, although the performance gains from extra data and multi-domain learning in the sequential framework are effective, it is significantly less than the parallel framework, which emphasizes the importance and necessity of the parallel framework.

\noindent{\textbf{Efficiency analysis of training and testing phases.}}
 We also provide efficiency analysis of training and testing phases between CENet and its variants. In training phase, we compare the training time of CENet with and without extra data in Table~\ref{tab:train_time_ablation}. It can be seen that introducing extra datasets significantly increases the training time. Specifically, training with only Night600 takes about 34 hours, while adding synthetic data extends it to 87 hours. Similarly, training with only RGBT201 takes about 3 hours, but with extra synthetic data, it extends to 72 hours. Notably, due to training on a public server, the training time is also related to the number of programs running on the server.
\begin{table}[ht]
\centering
\setlength{\tabcolsep}{1.5mm}
\caption{Ablation study of model training time before and after introduction of extra data}
\renewcommand\arraystretch{1.5}
\resizebox{0.88\linewidth}{!}{
\begin{tabular}{cccc}
\toprule
Training time & CENet   & CENet w/ Syn\_dark   & Platform                   \\ \hline
Night600     & 34 hours & 87 hours & \multirow{2}{*}{RTX3090*1} \\
RGBNT201    & 3 hours & 72 hours &                            \\ \bottomrule
\end{tabular}}
\label{tab:train_time_ablation}
\end{table}

In testing phase, we present the inference time comparison between with and without the relighting branch in Table~\ref{tab:test_time_ablation}. It can be seen that remove relighting branch saves 14 seconds and 33 seconds in night600 and Syn\_drak test datasets, respectively. Notably, the Syn\_dark test set is larger than the Night600 test set, leading to greater time savings for larger test sets.
\begin{table}[ht]
\centering
\setlength{\tabcolsep}{1.5mm}
\caption{Ablation Study of time savings before and after removing the relighting branch}
\renewcommand\arraystretch{1.5}
\resizebox{0.88\linewidth}{!}{
\begin{tabular}{cccc}
\toprule
Inference time & Night600   & Syn\_dark   & Platform                   \\ \hline
w/ relighting     & 85 seconds & 181 seconds & \multirow{2}{*}{RTX3090*1} \\
w/o relighting    & 71 seconds & 148 seconds &                            \\ \bottomrule
\end{tabular}}
\label{tab:test_time_ablation}
\end{table}

\subsection{Visualization}

\begin{figure}[h]
    \centering
    \begin{tabular}[b]{c}
    \includegraphics[width=0.45\textwidth,height=3.2cm]
    {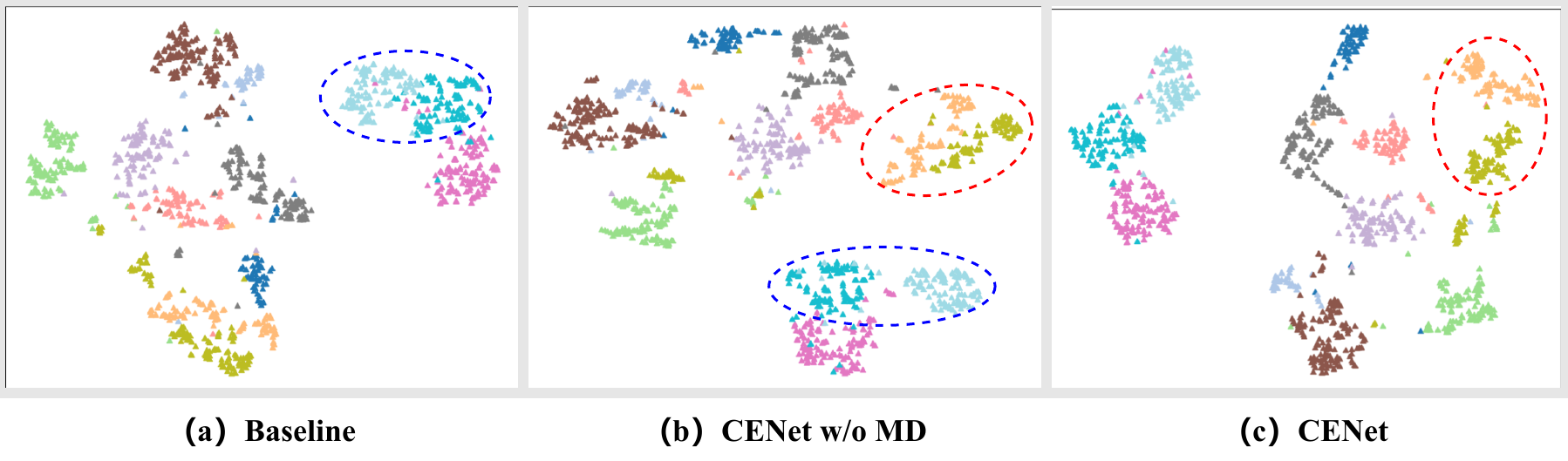}
    \end{tabular} 
    \caption{The t-SNE~\cite{tSNE_hinton} visualization of features on \textit{Night600} dataset.}
    \label{figure:tcts_tsne}
\end{figure}

\noindent{\bf Visualization of feature distribution.} To visualize the distribution of sample features, we randomly select 12 different samples from the test set and extract their features using three models: baseline, CENet w/o MD, and CENet. Then, we employed the t-SNE technique~\cite{tSNE_hinton} to project the high-dimensional features into a two-dimensional space. Figure~\ref{figure:tcts_tsne} (a) illustrates the representation distribution of these sample features by the baseline model, where different colored dots represent distinct identities. It can be observed that the sample features of different individuals within the blue dashed line region are difficult to distinguish. In contrast, CENet w/o MD demonstrates improved inter-class separation and intra-class cohesion, enhancing the overall feature distribution, as depicted in Figure~\ref{figure:tcts_tsne} (b). This demonstrates the effectiveness of the proposed network. Furthermore, the introduction of multi-domain learning in CENet further enhances intra-class compactness, leading to improved discriminability of hard samples, as highlighted by the red dashed region in Figure~\ref{figure:tcts_tsne} (c). This confirms the effectiveness of multi-domain learning in enhancing the feature representation for challenging samples.


\begin{figure}[h]
    \centering
    \begin{tabular}[b]{c}
    \includegraphics[scale=0.45]{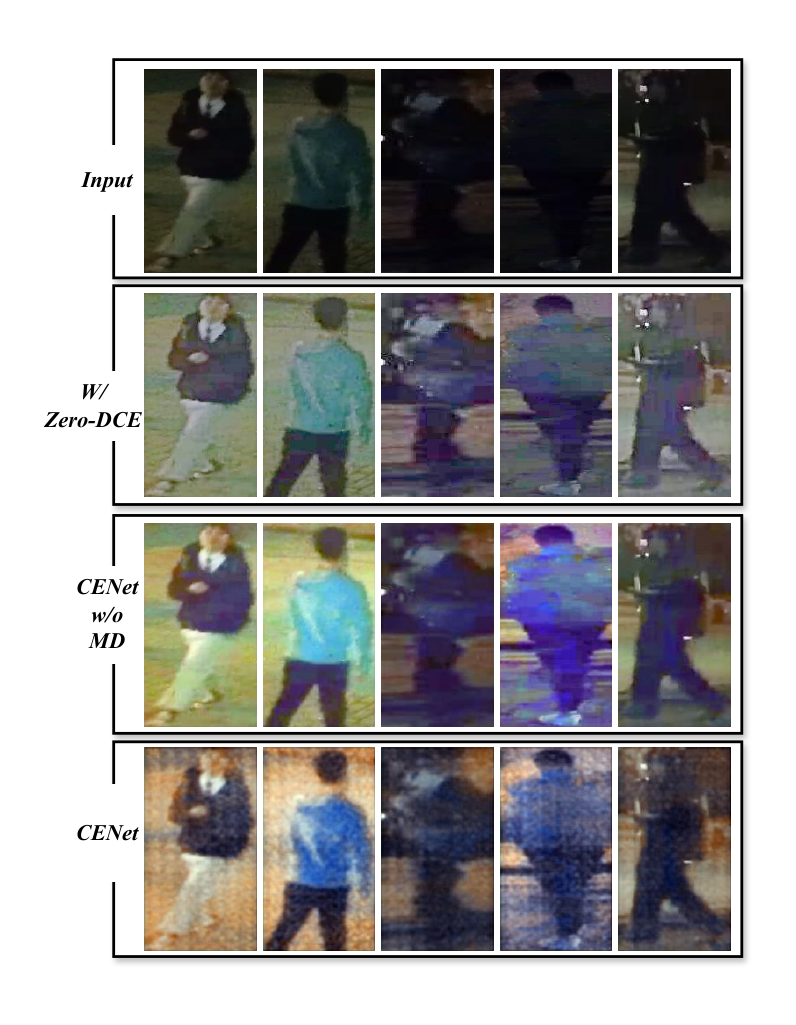}
    \end{tabular} 
    \caption{Visualization of relighting images from three models, including existing advanced relighting method Zero-DCE~\cite{dce++tpami}, CENet without multi-domain learning and CENet with multi-domain learning}
    \label{figure:enhance_imgs}
\end{figure}
\noindent{\bf Visualization of relighting images.} We present the original nighttime images, the images relighted by the advanced relighting algorithm Zero-DCE~\cite{dce++tpami}, and output results of the relighting branch based on the “CENet w/o MD” and the “CENet” in Figure~\ref{figure:enhance_imgs}. It can be seen that the enhancement quality of “W/ Zero-DCE” is relatively low, especially in terms of brightness and color. It also can be observed that the “CENet w/o MD” effectively enhances the brightness of low-light images, indicating the significance of increasing image brightness for improved nighttime ReID performance. However, the unsupervised learning approach in the relighting branch limits color shifting and noise smoothing, as illustrated in the second row of Figure~\ref{figure:enhance_imgs}. In contrast, CENet incorporates a supervised relighting learning process using synthetic data. CENet not only improves the image brightness but also enhances the color quality and smoothness of noise, leading to better performance. This highlights the importance of enhancing the color and noise quality of nighttime images for nighttime ReID performance.
However, due to the influence of noise, the overall image quality still has more room for improvement.

\begin{figure*}[t]
    \centering
    \begin{tabular}[b]{c}
    \includegraphics[width=1\textwidth]{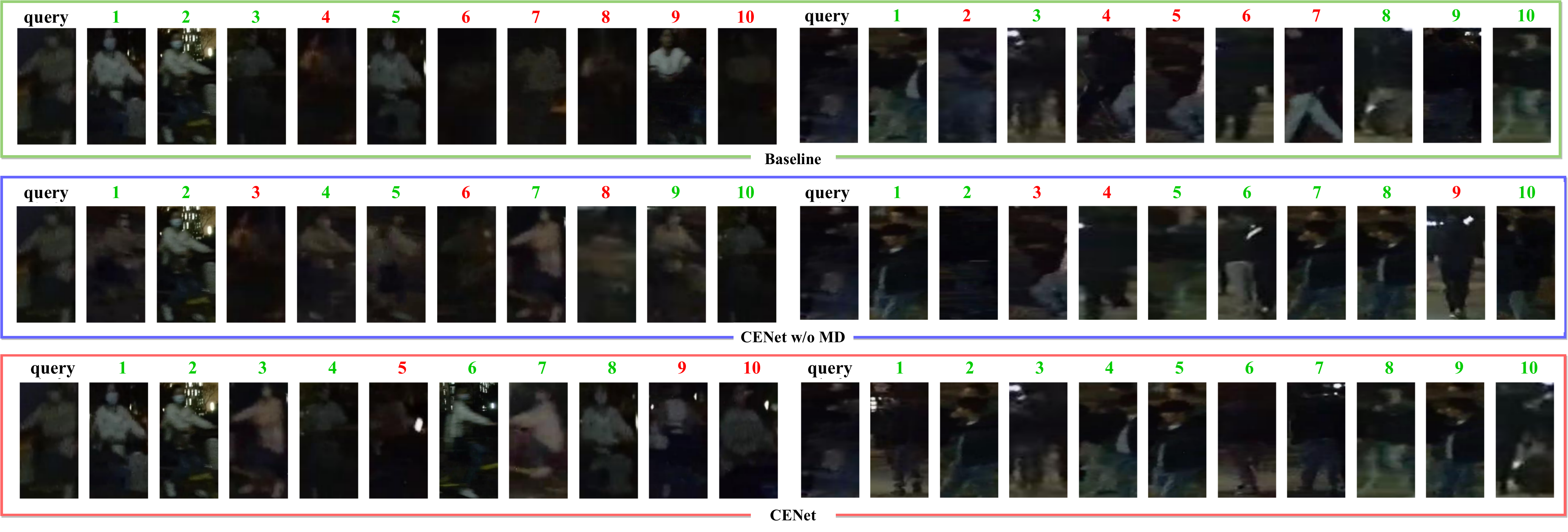}
    \end{tabular} 
    \caption{Visualization of sample ReID results from three models.}
    \label{figure:ranklisk_imgs}
\end{figure*}

\noindent{\bf Visualization of sample ReID results.} We also visualize the recognition results of a selection of representative samples to verify the effectiveness of the proposed method. In Figure~\ref{figure:ranklisk_imgs}, we present the recognition results of two examples under three models: “Baseline”, “CENet w/o MD” and “CENet”, respectively. It can be seen that the “Baseline” model mismatches a large number of very low light samples due to the difficulty in distinguishing them. However, in the results of “CENet w/o MD” model, the matching accuracy is effectively improved, even for some low-light samples. In “CENet”, the matching results are further improved, which shows the effectiveness of the proposed method. In addition, it can be observed in the demonstrated examples that the same person has obvious color shift problems in the night scene, which brings certain challenges to the conventional ReID method. However, our method can effectively avoid its influence, which indicates the necessity of studying the nighttime ReID algorithm.

\section{Conclusion}

We present a novel solution to address the challenge of person ReID in nighttime scenes. Our method leverages a parallel network architecture to simultaneously perform image relighting and person ReID, effectively avoiding the impact of relighting image quality on the ReID task. To achieve effective enhancement of ReID features from the relighting task, we design multilevel feature interaction based on parameter-sharing and feature distillation, which enhance both low-level and high-level features in ReID. In addition, we design a multi-domain learning scheme that learns more robust feature representations from data across different domains, significantly improving the model's performance in real night scenes. In future work, we aim to explore unsupervised learning schemes for nighttime ReID, which will be able to leverage larger-scale unlabeled real nighttime data, thereby further improving the performance of nighttime ReID.

\bibliographystyle{IEEEtran}
\bibliography{main}
 




\vfill

\end{document}